\documentclass[conference]{IEEEtran} 
\usepackage{graphicx} 
\usepackage{amsmath} 
\usepackage{cite} 
\usepackage{amsmath, amssymb, amsfonts}
\usepackage{hyperref}
\usepackage{subcaption}

\begin{document} 

\title{Generative Adversarial Synthesis and Deep Feature Discrimination of Brain Tumor MRI Images} 

\author{
    Md Sumon Ali$^{1}$\,[0009-0009-2749-9335],
    Muzammil Behzad$^{1,2,*}$\,[0000-0003-3693-4596]\\[4pt]
    $^{1}$King Fahd University of Petroleum and Minerals, Saudi Arabia\\
    $^{2}$SDAIA-KFUPM Joint Research Center for Artificial Intelligence, Saudi Arabia\\[4pt]
    Email: \url{g202320610@kfupm.edu.sa} and \url{muzammil.behzad@kfupm.edu.sa}
}

\maketitle 

\begin{abstract} 
Compared to traditional methods, Deep Learning (DL) becomes a key technology for computer vision tasks. Synthetic data generation is an interesting use case for DL, especially in the field of medical imaging such as Magnetic Resonance Imaging (MRI). The need for this task since the original MRI data is limited. The generation of realistic medical images is completely difficult and challenging. Generative Adversarial Networks (GANs) are useful for creating synthetic medical images. 
In this paper, we propose a DL based methodology for creating synthetic MRI data using the Deep Convolutional Generative Adversarial Network (DC-GAN) to address the problem of limited data. We also employ a Convolutional Neural Network (CNN) classifier to classify the brain tumor using synthetic data and real MRI data. CNN is used to evaluate the quality and utility of the synthetic images. The classification result demonstrates comparable performance on real and synthetic images, which validates the effectiveness of GAN-generated images for downstream tasks. \def\thefootnote{}\footnotetext{$^*$ indicates corresponding author.}
\end{abstract} 


\section{Introduction} 
\subsection{Background and Significance} 

Magnetic Resonance Imaging (MRI) is widely used noninvasive imaging method in medicine for diagnosing health issues and preparing treatment strategies. MRI is highly effective at delivering exceptional contrast, making it particularly valuable for evaluating brain disorders, cardiovascular conditions, and monitoring the progress of treatments. However, the extended acquisition time, which can be affected by hardware constraints or patient conditions, often leads to discomfort and may hinder urgent medical evaluation. In some cases the original MRI images is also limited \cite{ref15}. In this case, we need to generate synthetic images. In addition to traditional methods\cite{ref16}, Convolutional Neural Networks (CNN) have recently revolutionized medical image analysis\cite{ref17}, especially in the segmentation of brain MRIs\cite{ref18}.Training a CNN demands a significant volume of medical data, which is challenging to collect\cite{ref19}.

    \begin{figure}[htb]
    \centering
    \includegraphics[width=0.9\linewidth]{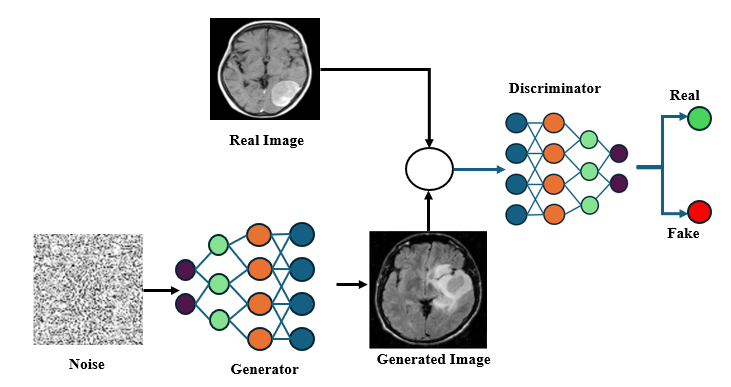}
    \caption{Overview of the image generation process using noise and real MRI image.}  
    \label{fig:overview}  
    \end{figure}

\subsection{Challenges in Current Techniques} 

Deep Learning (DL) has become a key technology in Computer Vision due to its exceptional performance, surpassing traditional methods. It is a Machine Learning (ML) strategy that supports the computer system to learn by itself through examples. Deep neural network is highly depends on the size of the datasets. The efficiency of DL models can be improved by training them on large volumes of data. In practice fields such as medical imaging, building a large dataset is difficult because of stringent medical image standards and strict patient privacy laws\cite{ref14}. Recently, General Adversarial Networks (GANs) have demonstrated significant potential in generating synthetic MRI images in Fig. \ref{fig:overview}, offering a promising approach for data augmentation and enable privacy-preserving research\cite{ref13}.

In this paper, we propose a Deep Convolutional GAN (DC-GAN) to generate synthetic MRI images employing a small-sized dataset. The generated images are used to train and evaluate a CNN-based tumor classifier alongside real MRI data. We want to train and test a CNN model to check the viability of using generated MRI images and real images to compare its performance against real MRI datasets.
The overall study consists of two main parts: (1) a synthetic image generation system using the DC-GAN, which is trained on actual brain MRI scans, and (2) a CNN classifier that is train and test using a combination of real and artificially generated datasets. The DC-GAN model consists of two key networks: a generator and a discriminator. The generator is designed to produce realistic MRI images, while discriminator is responsible for distinguishing between original and synthetic images\cite{ref32}. The final part consists of CNN classifier. The classifier is a CNN-based DL model that is trained on combination of both real and synthetic dataset and finally test it\cite{ref33}. This approach is used to evaluate the use of synthetic data affects classification performance.

\subsection{Problem Statement} 
Medical imaging, particularly MRI, plays a vital role in clinical diagnosis and treatment planning due to its unparalleled ability to provide detailed soft tissue contrast\cite{ref36}. However, its widespread application is limited by several challenges, such as the lack of annotated datasets, expensive acquisition process, and extended scanning duration\cite{ref35}.  Some challenges to accepting MRI as a method of healthcare have been reported, including technological, regulatory, and economical challenges\cite{ref34}. Traditional method like atlas-based registration are not only resource-intensive but also prone to inaccuracies\cite{ref20}. CNN-based methods, widely employed in medical image segmentation, frequently struggle with managing 3D spatial relationships and tend to experience performance plateaus as the depth the network grows\cite{ref19}. To overcome these limitation, the proposed approach utilizes a GAN-based synthetic image generation model that integrated Deep Convolutional GANs (DC-GANs) to generate synthetic brain MRI images for brain tumor classification.

\subsection{Scope of Study} 

This research focuses on generating and evaluating synthetic brain MRI images using DC-GAN and employing these images for medical image classification. The study aims to improve data augmentation techniques by leveraging GAN-generated medical images, with the goal of improving DL models for brain tumor detection. This study improves training a DC-GAN model to generate realistic 128x128 MRI images, assessing the quality if these synthetic images, and investigating their impact of classification performance when compared to real MRI images. The primary goal is to determine whether GAN-generated images can serve as a reliable alternative to real medical datasets for AI-based diagnosis and medical research application\cite{ref13}.

The rest of the paper is organized as: section \ref{related} describes the literature review and section \ref{method} outlines the methodology, while section \ref{experiment} describes the experiment, which are discussed in section \ref{discussion}; finally, section \ref{conclusion} conclude the paper, and put forward some future directions.

\section{Literature Review}
\label{related}

\subsection{Existing Techniques and Related Work}

Traditional data augmentation methods such as intensity and geometric transformations often fail to generate sufficiently diverse and correlated data samples, especially for medical imaging tasks where preserving anatomical accuracy is essential \cite{ref14,ref38}. To address these limitations, researchers have explored advanced deep generative approaches capable of producing high quality synthetic data \cite{ref37}. 

GANs are popular technology for making realistic synthetic images, videos, and more. They work by having two AI models compete against each other: one (Generator) tries to make fakes, and the other tries to identify the fakes images. This competition teaches the system to create convincing results \cite{ref21}. Several variants, such as Deep Convolutional GAN (DC-GAN) and Wasserstein GAN (WGAN), have been proposed to enhance stability and improve image realism. For instance, DCGANs have been employed to generate synthetic Positron Emission Tomography (PET) images \cite{ref26} and realistic retinal vessel segmentation images \cite{ref27}. Similarly, WGANs and multi GAN architectures have been applied to synthesize brain tumor MRI scans and computed tomography (CT) images, showing promise in augmenting limited medical datasets \cite{ref20,ref28}. 

Researchers have also used GANs to create medical images, like brain scans. This helps create more varied training examples to better educate doctors \cite{ref25}. In parallel, DL models, particularly CNN, have proven highly successful in image classification but remain dependent on large and well-annotated datasets. GAN-based augmentation has been shown to alleviate this limitation by producing realistic synthetic samples that improve model generalization \cite{ref22,ref23}. 

Recent studies are mixing GANs with other types of deep learning to fix common problems. This helps create more reliable and anatomically correct images \cite{ref13}. Additionally, style transfer techniques have been incorporated to enhance the realism of generated medical images \cite{ref28}. Collectively, these studies have established GAN-based augmentation as a critical tool for medical imaging research, with demonstrated benefits in MRI, CT, and other modalities \cite{ref40,ref42}.

\subsection{Limitations in Existing Approaches}

However, there is still a significant problem with GANs called mode collapse. This is when the AI becomes stuck and creates only a few, very similar images instead of a wide variety. This lack of diversity makes the data less useful for doctors \cite{ref14,ref29}. Furthermore, while GAN-generated images may appear visually realistic, they often lack essential anatomical and pathological details necessary for accurate diagnosis \cite{ref13}. The adversarial training process is also notoriously unstable, requiring careful tuning of hyperparameter to achieve convergence and avoid artifacts \cite{ref28}. 
Other technical challenges include vanishing gradients, non convergence, and the requirement for large, well annotated datasets resources that are often limited due to privacy concerns and data scarcity \cite{ref12,ref20}. 

To fix these issues, researchers have created better versions of GANs, like one called DC-GAN, which produces more stable and reliable images. Hybrid methods that combine synthetic and real MRI data during classifier training have also been recommended to mitigate the effects of artifacts and overfitting. The proposed DC-GAN-based system in this study seeks to overcome existing limitations by generating realistic brain MRI images and integrating them into a CNN based classification framework for tumor detection. 

A comprehensive comparison of the reviewed methodologies, along with their limitations, is presented in TABLE \ref{tab:relatedcompartson}. Based on these findings, our proposed DC-GAN model aims to enhance the realism, diversity, and diagnostic quality of synthetic brain MRI images, ultimately improving performance in brain tumor classification tasks.

\begin{table*}[t]
\centering
\caption{Comparison of GAN-based Methodologies and Their Limitations}
\begin{tabular}{|l|p{8cm}|}
\hline
\textbf{Methodology Name} & \textbf{Limitation} \\
\hline
GAN \cite{ref21} & Suffers from mode collapse, unstable training \\
\hline
DCGAN \cite{ref22} & Blurry outputs, limited detail \\
\hline
Improved GAN \cite{ref23} & Focuses on multi-class generation, lacks anatomical precision \\
\hline
Likelihood-based models \cite{ref24} & Blurry results due to noise and incomplete reconstruction \\
\hline
GAN for brain tumor MRI \cite{ref25} & May generate realistic images but lacks anatomical fidelity \\
\hline
DCGAN for PET \cite{ref26} & Performance limited by dataset quality \\
\hline
CNN + GAN for retinal images \cite{ref27} & Limited diversity, artifacts in outputs \\
\hline
GAN for brain synCT \cite{ref20} & GAN suffers from mode collapse; anatomical features lost \\
\hline
Aggregated GAN (DCGAN + WGAN) \cite{ref28} & Complex training, hyperparameter tuning challenges \\
\hline
Deep learning for MRI enhancement \cite{ref42} & Needs large datasets, computationally expensive \\
\hline
Intensity-based augmentation \cite{ref38} & Low variability, fails to preserve medical detail \\
\hline
Advanced GANs for diversity \cite{ref39} & Still struggle with high-res and anatomically accurate images \\
\hline
GANs for synthetic MRIs \cite{ref40} & Privacy issues, limited annotations \\
\hline
Multi-sequence MRI via GAN \cite{ref13} & Suffers mode collapse, reduced diversity \\
\hline
Basic augmentation \& GAN issues \cite{ref14} & Mode collapse, insufficient data complexity \\
\hline
GAN-based tumor classifier \cite{ref29} & Overfitting due to low diversity in synthetic data \\
\hline
One-to-one mapping GAN \cite{ref30} & Lacks protocol/scanner variability \\
\hline
GAN gradient problems \cite{ref12} & Vanishing gradients, long training cycles \\
\hline
Existing GAN challenges \cite{ref6}, \cite{ref7} & Mode collapse, data bias, vanishing gradients \\
\hline
Domain adaptation with GANs \cite{ref8}, \cite{ref9} & Need for better bridging of synthetic-real domain gap \\
\hline
\end{tabular}
\label{tab:relatedcompartson}
\end{table*}

\section{Proposed Methodology} 
\label{method}


The main study of this work is to develop a DC-GAN, which is used to generate synthetic brain MRI images and then a CNN used for training with generated images and real images to classify both the images. The DC-GAN consists of a Generator, which learns to produce realistic images from random noise, and a Discriminator, which distinguishes between real and artificially generated images\cite{ref7}.
Both networks are trained in a min-max game based on the following objective function (loss function) of GANs:

\begin{align}
\min_G \max_D V(D, G) =\; & \mathbb{E}_{x \sim p_{\text{data}}(x)} \left[ \log D(x) \right] \notag \\
                         +\; & \mathbb{E}_{z \sim p_z(z)} \left[ \log \left(1 - D(G(z)) \right) \right] \label{eq:gan_loss}
\end{align}

where:

\begin{itemize}
    \item \( p_{\text{data}}(x) \): Real data distribution (e.g., MRI images),
    \item \( p_z(z) \): Prior distribution on input noise variables (latent space),
    \item \( G(z) \): Output of the generator (fake MRI image),
    \item \( D(x) \): Discriminator's estimated probability that \( x \) is real.
\end{itemize}

The CNN classifier is trained on both real and synthetic MRI images to classify the tumor, how accurately the synthetic data mimics real-world data.

\subsection{Proposed Enhancements} 

To address these issues, which were discussed in the literature review section, several improvements have been introduced. It starts training with low-resolution images and progressively increases the depth and resolution of the network. To ensure stable training and avoid exploding gradients, we apply spectral normalization in the Discriminator. Additionally, inspired by the Auxiliary Classifier Generative Adversarial Network (ACGAN) framework\cite{ref31}, we enhance the Discriminator with an auxiliary classifier, allowing it to not only differentiate between real and synthetic images but also classify tumor types.To make the AI models stronger and reliable, researchers use new methods to alter both real and AI generated images by stretching them and changing their brightness. This creates more varied examples for training. Finally, a hybrid classification model is adopted, wherein the CNN classifier is initially trained on real and synthetic MRI images and then test it, again both real and synthetic MRI images to identify there is no major difference between real and synthetic images. These enhancements are consistent with the latest progress in medical image synthesis.

\subsection{Algorithm and Implementation} 
The described methodology follows a three-stage process:
\subsubsection*{Dataset and Preprocessing}

For this project, we used the Brain tumor MRI Images for Brain Tumor Detection datasets from Kaggle, which includes 3000 grayscale MRI scans \cite{ref43,ref44}. These images are divided into two categories: 1500 labeled "yes" (showing tumors) and 1500 labeled "no" (no tumors). Since the original scans varied in size and quality, we preprocessed them to ensure consistency.
Each image was resized to 128×128 pixels, converted to grayscale, and normalized to a pixel range of [-1, 1] matching the output range of our GAN’s generator. Finally, we compiled the processed images into a NumPy array, creating a clean, standardized dataset ready for training. This step was crucial for maintaining stability and improving the performance of our DC-GAN model.

\subsubsection*{GAN Training}
A DC-GAN architecture was employed to generate synthetic brain MRI images from a random noise vector of size 400. The Generator comprises a deep convolutional design with a dense input layer projecting to a feature map of size $32 \times 32 \times 256$, followed by a sequence of transposed convolutional layers that progressively upscale the spatial resolution to $128 \times 128$. Each convolutional block uses \texttt{LeakyReLU} activations to enhance gradient flow and prevent vanishing gradients. The final convolutional layer applies a single output channel with a \texttt{tanh} activation to produce normalized grayscale MRI images. The Discriminator consists of four convolutional layers that progressively downsample the input from $128 \times 128$ to $16 \times 16$ feature maps, each followed by \texttt{LeakyReLU} activations and dropout regularization to mitigate overfitting. The final flattened feature map is passed through a dense layer with a sigmoid output to classify inputs as real or synthetic. The network was trained for 10 epochs with 3{,}750 steps per epoch and a batch size of 4, using the Adam optimizer to ensure stable convergence between the generator and discriminator.

\subsubsection*{CNN Classification}
We built a CNN model to find brain tumors. It was trained to check both real MRI images and synthetic ones created by the DC-GAN model.
Each block employed multiple \texttt{Conv2D} layers with \texttt{ReLU} activation, \texttt{Batch Normalization}, and \texttt{MaxPooling2D}, combined with dropout regularization to prevent overfitting. The final block used \texttt{GlobalAveragePooling2D} to reduce feature dimensions, followed by a dense layer with 1024 neurons and an output layer with a \texttt{sigmoid} activation for binary classification. The model, optimized using Adam with a learning rate of 0.0005, incorporated early stopping, learning rate reduction, and checkpoint callbacks to ensure stable convergence and prevent overfitting during 200 training epochs.

The algorithm follows:
\begin{itemize}
    \item Load MRI datasets and apply preprocessing.
    \item Train using alternating Generator-Discriminator updates.
    \item Save synthetic images at specific epochs.
    \item Merge real and synthetic images and augmented it.
    \item Divide the merged images for train and test.
    \item Train CNN classifier on real and synthetic merged MRI train images.
    \item Test the CNN classifier and classify the tumor images from the merged test data.
    \item Compare performance metrics between real and synthetic images.
\end{itemize}

\subsection{Loss Function and Optimization} 
For GAN training, Binary Cross-Entropy Loss (BCE Loss) is used to optimize both Generator and Discriminator\cite{ref10}\cite{ref11}. Given a real image \( x \) and generated image \( G(z) \), the Discriminator loss \( L_D \) is:

\begin{equation}
L_D = -\mathbb{E}[\log D(x)] - \mathbb{E}[\log(1 - D(G(z)))]
\end{equation}

The Generator aims to fool the Discriminator, optimizing the loss:

\begin{equation}
L_G = -\mathbb{E}[\log D(G(z))]
\end{equation}

Adam optimizer is used with learning rate 0.0002 and momentum parameters (\(\beta_1=0.5\), \(\beta_2=0.999\)), ensuring stable updates. For CNN classification, Cross-Entropy Loss is applied:

\begin{equation}
L_{CE} = - \sum y_i \log \hat{y}_i
\end{equation}

where \( y_i \) is the ground truth and \( \hat{y}_i \) is the predicted probability. 

where \( y_i \) is the ground truth and \( \hat{y}_i \) is the predicted probability. The CNN is optimized using Adam with dropout regularization, preventing overfitting\cite{ref41}. The training progress is monitored through loss curves and classification accuracy. These optimization techniques align with state of the art practices in medical image synthesis and classification .

\section{Experimental Design and Evaluation} 
\label{experiment}


\subsection{Experiment Setup} 

This study had two main phases: first, generating synthetic brain MRI scans using a specialized DC-GAN, and second, training and testing another CNN model to detect tumors in those images. We started with a dataset of real brain scans 1500 with tumors and 1500 without then resized them to 128x128 pixels and adjusted their brightness values to help the model learn better. We generated 400 yes images and then merged them with the original images. To balance the dataset for CNN classification we augmented no images up to 400. The DC-GAN worked like an art forger and detective playing against each other: one part created fake MRI images from random noise, while another part tried to spot the fakes. They trained together in repeated cycles using an efficient learning method called Adam optimizer, gradually improving until the fake scans looked convincing.\\

For the tumor detection phase, we mixed real tumor scans with the generated ones to create a more balanced dataset, then resized everything to 128x128 pixels for consistency. The tumor-detecting model used a layered structure with built-in safeguards (like dropout and L2 regularization) to prevent overfitting. We trained it on 80\% of the data while reserving 20\% for testing, artificially expanded the training set by flipping and rotating scans, and let it run for up to 200 iterations. To check its performance, we looked at accuracy trends, mistake patterns, and specialized scores (precision, recall, F1) that reveal how well it distinguishes tumors from healthy tissue.

    \begin{figure}[htb]
    \centering
    \includegraphics[width=0.9\linewidth]{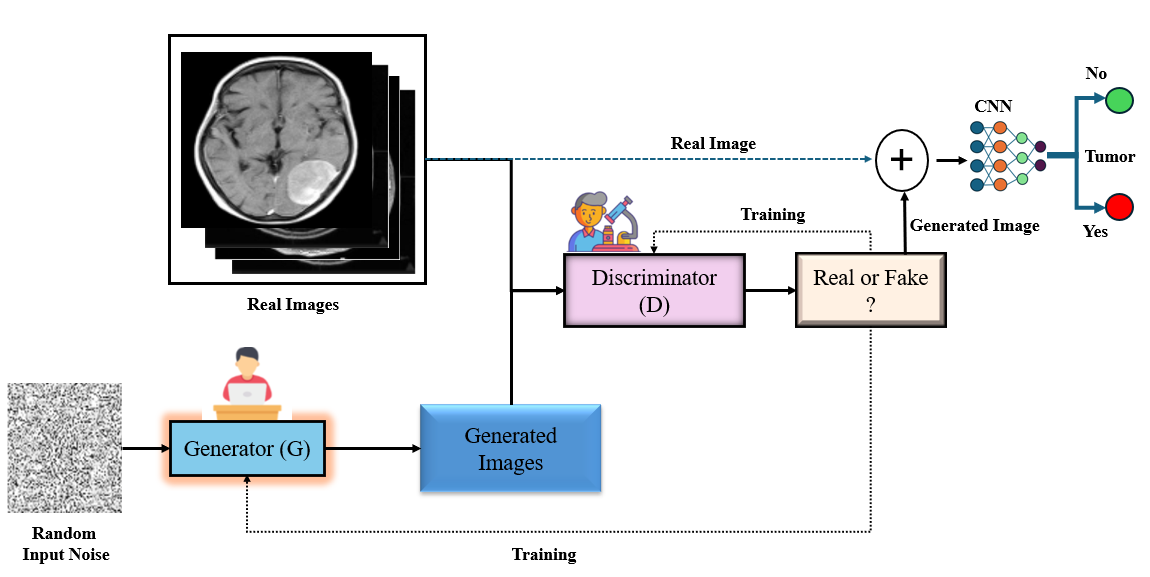}
    \caption{Architecture of the DC-GAN model for brain MRI image generation and Classification.}  %
    \label{fig:architecture}  
    \end{figure}

The two neural networks the generator and the discriminator used in a training process were like a constant competition which is shown in Fig. \ref{fig:architecture}. Starting with nothing but random noise, the generator slowly learned how to create more detailed and realistic outputs 128x128 grayscale images that looked almost identical to real brain scans. Meanwhile, the discriminator acted as a careful gatekeeper, carefully examining each image to spot whether it was a real MRI scan from our dataset or a synthetic one created by the generator.
The two networks took turns improving first, the discriminator got better at detecting fakes, then the generator countered by producing even more convincing images. Over time, this back and forth pushed both to new heights: the generator’s brain scans became incredibly original, while the discriminator sharpened its ability to tell real from fake with impressive accuracy.\\

To track the model’s progress, we focused on two key metrics: the generator loss and the discriminator loss. The generator’s loss told us how well its fake images were tricking the discriminator basically, how often the discriminator got fooled into thinking the synthetic scans were real. Over time, we saw this number steadily drop, a sign that the generator was getting better at creating realistic  brain scans. On the other hand, the discriminator’s loss measured how good it was at catching fakes. While we wanted this value to stay low, we noticed something interesting: if the discriminator got too good too quickly at the beginning, it actually slowed down the generator’s progress. \\

While the numbers gave us one way to track progress, we also used simple visual checks to see how the model was performing. After each training round, we generated sample brain scans from random noise and looked at them carefully to see how realistic they were.

    \begin{figure}[htb]
    \centering
    \includegraphics[width=0.9\linewidth]{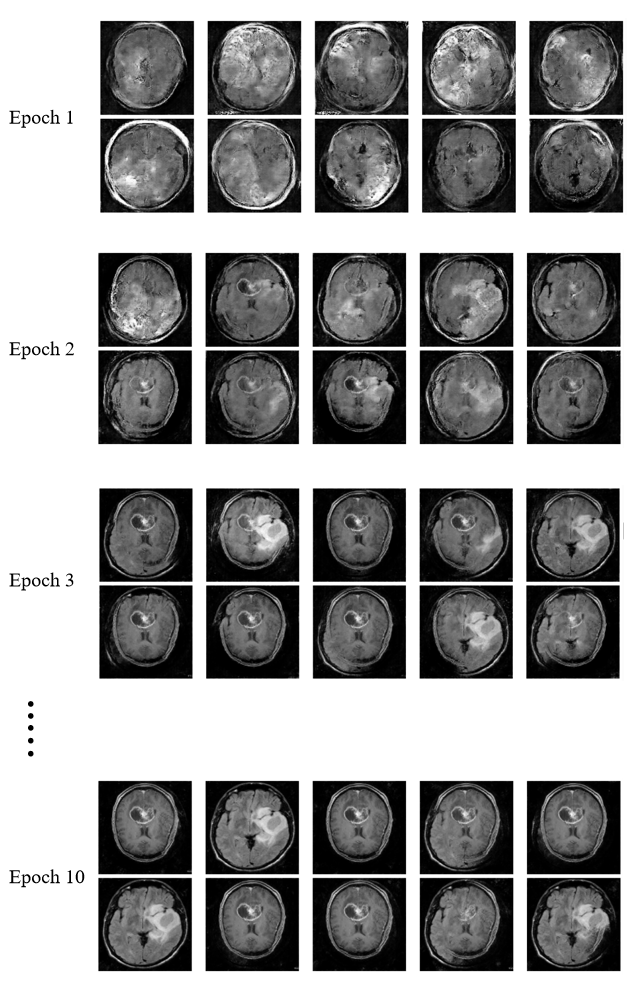}
    \caption{Progression of Brain Tumor MRI Image Generation Across Training Epochs.}  
    \label{fig:epoch}  %
\end{figure}

The first few attempts (epochs 1-3) didn't look great they were blurry, oddly shaped, and only somewhat resembled real MRIs. But as training continued, the improvements were really noticeable. By epochs 5-10, the model started generating scans with clear and sharp anatomical details and realistic looking tumors, complete with the subtle textures see in real patient images which is shown in Fig. \ref{fig:epoch}. These side by side comparisons were eye opening they clearly showed how much the model had improved. It clearly showed that, our model had actually learned to create convincing synthetic brain scans.\\

To build an intelligent tumor detection system, we enhanced our GAN setup by adding a CNN classifier that could analyze both real and generated MRI scans. First, we prepared our dataset 1500 scans with tumors and 1500 without tumors. We added 400 synthetic tumor images with original tumor images and got a total of 1900 tumor images. We augmented no tumor images 400 because without tumor has less amount of images so it was imbalanced. To remove imbalances we we augmented it. Then resizing them to 128×128 grayscale images and adjusting their contrast for consistency. Then, we split the data 80/20 for training and testing. The real magic happened when we combined forces: our GAN generated realistic synthetic scans to boost the dataset, while the CNN learned how to detect tumors in both real and synthetic images.  A promising diagnostic assistant that could one day help doctors identify tumors more efficiently.\\

For our tumor detection system, we built a simple CNN model perfect for classifying images into two categories (in our case, "tumor" or "no tumor"), which is also shown in Fig. \ref{fig:architecture}. The model started with convolutional layers, scanning the MRI images to pick out important patterns. These were followed by max pooling layers to simplify the information, kind of like summarizing key points. Then we flattened everything out and fed it through some dense layers, with a final sigmoid activation that gave us a clear yes/no prediction about tumor presence.\\

We trained it using binary cross entropy loss (basically measuring how often it guessed wrong) and used the Adam optimizer to help it learn efficiently. To evaluate performance, we tracked not just accuracy but also precision, recall, and F1-score on the test set. This approach not only validated our classification pipeline but also demonstrated the practical value of the GAN generated images particularly in addressing the challenge of limited real annotated medical data. The results showed that synthetic data could effectively augment our training dataset when real MRI scans weren't available.\\

\subsection{Performance Metrics}

In the first few epochs, the generated images were blurry and lacked clear structure, which is shown in Fig. \ref{fig:epoch}. 
By epochs 2 and 3, the images started to improve its structure. Now the faint outlines of features and small spots also showed, which is possibly brain tumors begin to appear. Although they are still still blurry and not very clear at that moment. 
In epochs 4 and 5, the images become clearer and more detailed and the quality of the images improves with an increase in the epochs.
The generated brain image shapes start to look clearer, and the fake brain tumor images look more real. They fit in with the rest of the images instead of looking like random spots.
After 6 epochs, the generated image quality is amazing. The images look like real MRI scans. At this stage, anyone can see the brain parts and identify the tumor. The model learns how real medical images are supposed to look.
Finally, by epoch 10, the results look quite good. The MRI brain tumor images look almost like real ones, with clear tumor details and realistic. \\
    
    \begin{figure}[htb]
    \centering
    \includegraphics[width=0.9\linewidth]{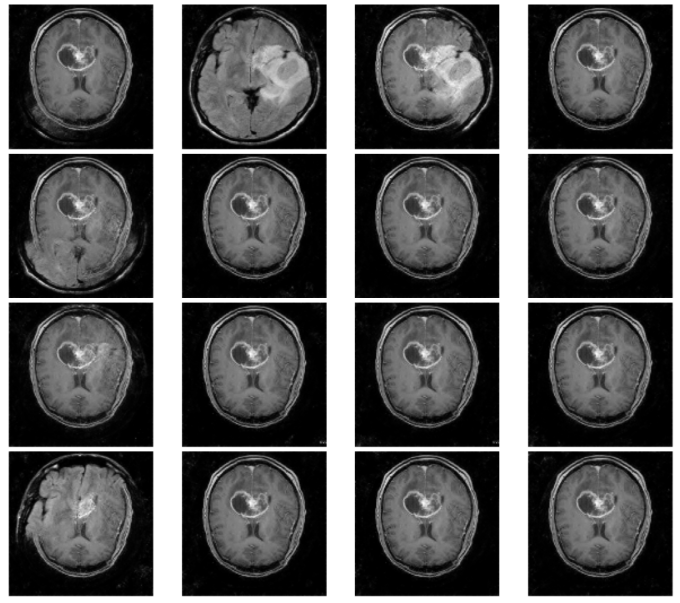}
    \caption{Generated Brain Tumor MRI Images After Final Training Epoch}  
    \label{fig:genearaed} 
\end{figure}

Fig. \ref{fig:genearaed} shows a grid of synthetic brain tumor MRI images which are made by the DC-GAN model. The grid shows how much the model has improved. Each image of the grid looks very real with clear brain details the folds of the brain and the shape of the brain tumors. Previous images were blurry and noisy, but now they have sharp textures and look like real medical scans with believable tumors.

    \begin{figure}[htb]
    \centering
    \includegraphics[width=0.9\linewidth]{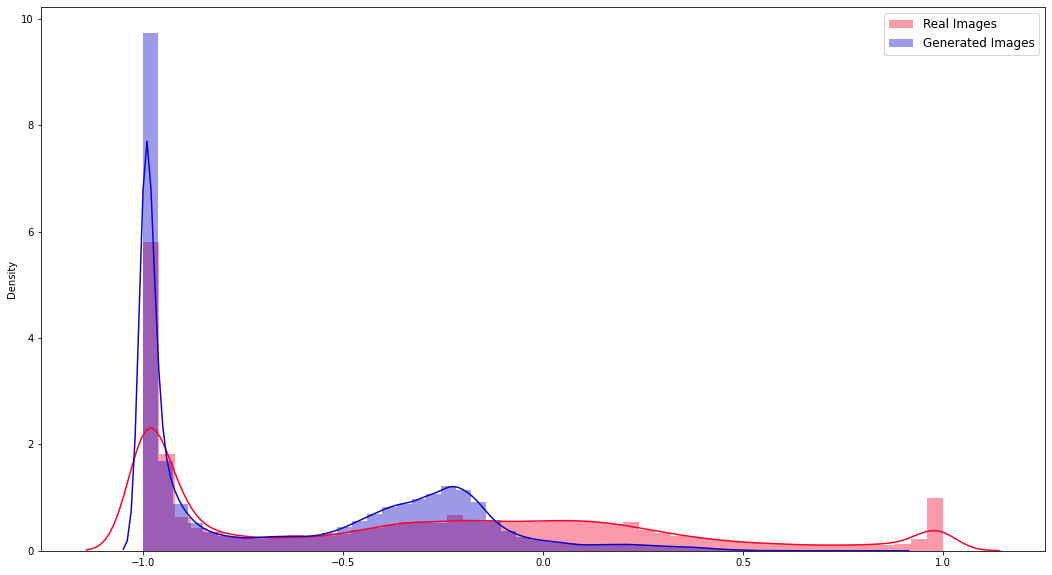}
    \caption{Distribution Comparison Between Real and Generated Brain MRI Images}  
    \label{fig:distribtuion} 
\end{figure}

The distribution plot in Fig. \ref{fig:distribtuion} shows something impressive. The red line (real MRIs) and the blue line (generated images) follow almost the same distribution. This indicates that our DC-GAN model not only creates images that look real, it also identifies the hidden patterns in real medical brain tumor scans.\\

\begin{figure}[htb]
    \centering

    \begin{subfigure}[b]{1.0\linewidth}
        \includegraphics[width=\linewidth]{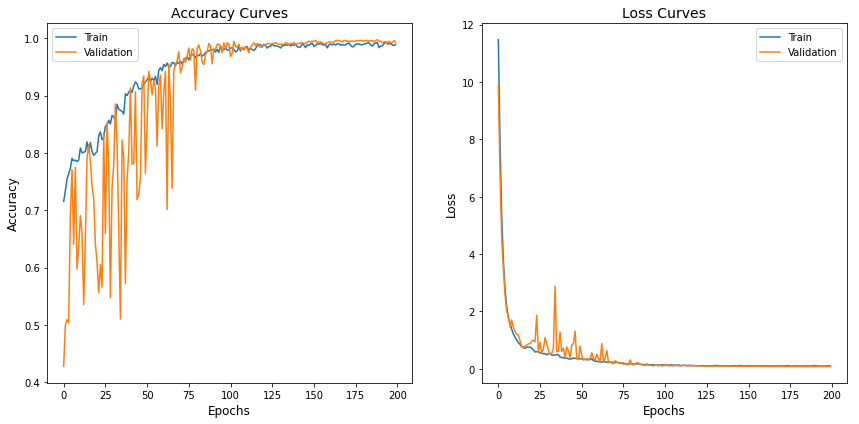}
        \caption{Accuracy and Loss over Epochs}
        \label{fig:fig6a}
    \end{subfigure}
    
    \vspace{0.5cm}  

    \begin{subfigure}[b]{0.7\linewidth}
        \includegraphics[width=\linewidth]{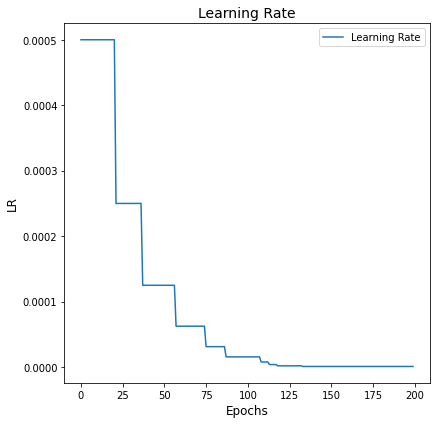}
        \caption{Learning rate over epochs}
        \label{fig:fig6b}
    \end{subfigure}

    \caption{Training and Testing Performance for CNN Architecture: Accuracy and Loss over Epochs}
    \label{fig:training}
\end{figure}

\begin{figure}[htb]
    \centering
    \includegraphics[width=0.7\linewidth]{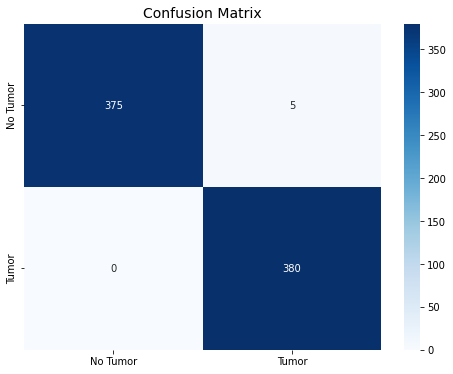}
    \caption{Confusion Matrix for Brain Tumor Classification Using Generated MRI Images}  
    \label{fig:confusion} 
\end{figure}

The training and validation accuracy and loss curves for CNN architecture are shown in Fig. \ref{fig:training}(a).
The model quickly achieved 90\% accuracy within the first 50 epochs. Fluctuation in the validation accuracy stated that the model struggled with the generated images. After some time when the model was perfectly learned, the training and validation accuracy curves were the same. Eventually, both training and validation accuracies stabilized near 99.7\%. The training loss went down smoothly over time, while the validation loss showed initial variability but later converged. This early fluctuation happened because the model was learning from a mix of different images, including augmented and DC-GAN-generated ones. Fig. \ref{fig:training}(b) presents the learning rate (LR) schedule over 200 training epochs. The training began with a high LR of 0.0005, which progressively reduced based on validation loss feedback through ReduceLROnPlateau callback. This adaptive learning strategy stabilized the training process, allowing the model to converge efficiently while avoiding oscillations or overfitting.\\
Fig. \ref{fig:confusion} presents the confusion matrix of the final model evaluated on the test dataset (20\%). The model achived only 5 false positive and 0 false negatives, which is particularly important in medical diagnostic. The high true positive and true negative counts indicate excellent classification ability. Synthetic images helped the model get better at finding tumors, which is shown in TABLE \ref{tab:classi}. The following metrics were computed from the test set (n=760). A precision of 1.00 for No tumor means every predicted negative case was correct. Recall of 1.00 for Tumor means all actual tumor cases were identified. A macro average F1-score of 0.99 indicated overall excellent performance across both classes.

\begin{table}[h!]
\centering
\caption{Classification Report for CNN Model}
\begin{tabular}{lcccc}
\hline
\textbf{Class} & \textbf{Precision} & \textbf{Recall} & \textbf{F1-Score} & \textbf{Support} \\
\hline
No Tumor       & 1.00               & 0.99            & 0.99              & 380               \\
Tumor          & 0.99               & 1.00            & 0.99              & 380               \\
\hline
Accuracy       & \multicolumn{3}{c}{0.99}            & 760               \\
\hline
Macro Avg      & 0.99               & 0.99            & 0.99              & 760               \\
Weighted Avg   & 0.99               & 0.99            & 0.99              & 760               \\
\hline
\end{tabular}
\label{tab:classi}
\end{table}


\begin{table}[h!]
\centering
\caption{Comparison of existing studies and proposed model for synthetic MRI generation and tumor classification.}
\label{tab:compartson}
\begin{tabular}{|p{1 cm}|p{1.0cm}|p{1.5cm}|p{1.5cm}|p{1.5cm}|}
\hline
\textbf{Study} & \textbf{Dataset} & \textbf{Model} & \textbf{Evaluation Metric(s)} & \textbf{Best Quantitative Results} \\
\hline
\textbf{Our Study } & Kaggle Brain MRI (3k images) & DCGAN, CNN & Accuracy, Precision, Recall, F1 & Accuracy = 99.7\%, F1 = 0.99 \\
\hline
Han et al.,\cite{ref13} & BraTS 2016 (HGG) & DCGAN vs WGAN & Visual Turing Test (Physician) & WGAN $\approx$ 53--55\% accuracy (chance = 50\%) \\
\hline
Mukherjee et al.,\cite{ref28} & Brain Tumor \& BraTS 2020 & AGGrGAN (DCGAN, WGAN, Style Transfer) & SSIM, PSNR, KL, SD, CNN Accuracy & SSIM = 0.83, PSNR = 23.7 dB, Accuracy = 94\% (VGG-19) \\
\hline
\end{tabular}
\label{tab:comparison3}
\end{table}

    \begin{figure}[htb]
    \centering
    \includegraphics[width=0.9\linewidth]{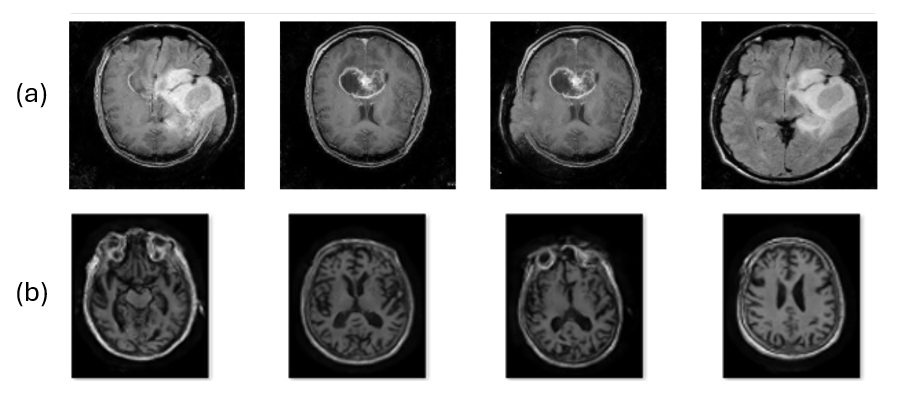}
    \caption{Comparison of synthetic brain MRI images generated by different methods.
(a) Images generated by our DC-GAN based model.
(b) Artificial MRI images from a prior GAN-based method.}  
    \label{fig:comparison_figure}  
\end{figure}


\section{Discussion}
\label{discussion}
The experimental findings from our proposed DC-GAN and CNN-based framework demonstrate significant advancements in synthetic brain MRI generation and tumor classification when compared with existing GAN-based approaches. As shown in TABLE \ref{tab:comparison3}, our model achieved an outstanding classification accuracy of 99.7\% and an F1-score of 0.99, outperforming prior studies such as Han \textit{et al.}~\cite{ref13} and Mukherjee \textit{et al.}~\cite{ref28}. While Han \textit{et al.} employed a Visual Turing Test to assess realism, their WGAN model achieved only (53-55)\% accuracy close to random guessing indicating that although visually plausible, the synthetic MRIs lacked sufficient diagnostic realism. In contrast, Mukherjee \textit{et al.} proposed an advanced AGGrGAN architecture combining multiple GAN variants with style transfer, achieving SSIM = 0.83, PSNR = 23.7 dB, and 94\% classification accuracy. Despite these improvements in visual fidelity and structural similarity, our approach demonstrated superior diagnostic performance, suggesting that realism alone is insufficient without a strong classification backbone capable of leveraging synthetic data effectively.

Fig. \ref{fig:comparison_figure} highlights qualitative differences between our generated MRIs and those from previous GAN based models. The synthetic images produced by our DC-GAN (Fig.~\ref{fig:comparison_figure}a) exhibit sharper textures, well defined brain structures, and more realistic tumor boundaries compared to earlier methods (Fig.~\ref{fig:comparison_figure}b). Additionally, the progressive learning behavior illustrated in Fig. \ref{fig:genearaed} and Fig. \ref{fig:training} confirms that the generator improved image quality substantially across epochs, while the CNN classifier achieved stable convergence without overfitting. These findings reinforce that high quality synthetic medical images can not only enhance data diversity but also significantly boost diagnostic model accuracy. Overall, the proposed framework demonstrates a robust and efficient strategy for addressing data scarcity in medical imaging, paving the way for reliable AI-assisted diagnostic systems that integrate both generative and discriminative deep learning paradigms.

\section{Conclusion and Future Work} 
\label{conclusion}

This study successfully demonstrates the capability of DC-GAN to generate realistic synthetic brain MRI images that can effectively augment limited medical datasets. The proposed approach enhanced the performance of a CNN based tumor classification model, achieving high accuracy and reliable tumor detection, as supported by the confusion matrix and detailed classification metrics. The GAN generated images showed progressive improvement across training epochs, producing clearer anatomical structures and more realistic tumor textures, which improved the classifiers learning ability and generalization performance.
However, several limitations remain that provide valuable directions for future research. Future work will focus on improving evaluation rigor by incorporating patient level data splits, external validation sets, and clinically meaningful statistical measures such as sensitivity, specificity, ROC-AUC, and confidence intervals. Additionally, quantitative image quality metrics including SSIM, PSNR, FID, and KID, along with blinded expert evaluations, will be employed to validate anatomical realism, while model configurations will be standardized by aligning GAN and CNN resolutions and unifying hyperparameters to enhance reproducibility.
Finally, dataset licensing, ethical use, and privacy aspects of synthetic medical data will be explicitly addressed, and the framework will be extended to higher resolution and multi institutional datasets to enhance its clinical applicability.
Overall, this research highlights the potential of GAN generated data as a powerful augmentation tool in medical imaging and sets the foundation for more rigorous, interpretable, and ethically grounded advancements in AI driven diagnostic systems.


\end{document}